\documentclass{article} 
\usepackage{drcredit_preprint,times}

\usepackage{amsmath,amsfonts,bm}

\def\eqref#1{equation~\ref{#1}}

\def\1{\bm{1}}

\DeclareMathAlphabet{\mathsfit}{\encodingdefault}{\sfdefault}{m}{sl}
\SetMathAlphabet{\mathsfit}{bold}{\encodingdefault}{\sfdefault}{bx}{n}

\usepackage{hyperref}
\usepackage{url}
\usepackage{graphicx}
\usepackage{amssymb}
\usepackage{booktabs}
\usepackage{multirow}
\usepackage{colortbl}
\usepackage{pifont}
\usepackage{float}
\usepackage{wrapfig}

\title{Dr.Credit: Rubric-Grounded Process \\
Credit Assignment for Deep Research Agents}

\author{%
  \textbf{Yingjian Zhu}\textsuperscript{1,2,3}\thanks{Work done during an internship at Alibaba Token Hub, Alibaba Group.}\quad
  \textbf{Zhenyi Wang}\textsuperscript{3}\thanks{Co-corresponding authors: Kun Ding and Zhenyi Wang.}\quad
  \textbf{Jiaxin Guo}\textsuperscript{3,4}\\[3pt]
  \textbf{Kun Ding}\textsuperscript{2}\footnotemark[2]\quad
  \textbf{Ying Wang}\textsuperscript{2}\quad
  \textbf{Shen Huang}\textsuperscript{3}\\[3pt]
  \textbf{Xunjie Zhu}\textsuperscript{3}\quad
  \textbf{Pengjun Xie}\textsuperscript{3}\quad
  \textbf{Shiming Xiang}\textsuperscript{2}\\[7pt]
  {\normalfont\small\textsuperscript{1}School of Artificial Intelligence,}\\
  {\normalfont\small University of Chinese Academy of Sciences}\\[3pt]
  {\normalfont\small\textsuperscript{2}State Key Laboratory of Multimodal Artificial Intelligence Systems (MAIS),}\\
  {\normalfont\small Institute of Automation, Chinese Academy of Sciences}\\[3pt]
  {\normalfont\small\textsuperscript{3}Alibaba Token Hub, Alibaba Group}\quad
  {\normalfont\small\textsuperscript{4}Peking University}%
}

\input{appendix/prompt_setup}

\hypersetup{
  pdftitle={Dr.Credit: Rubric-Grounded Process Credit Assignment for Deep Research Agents},
  pdfauthor={Yingjian Zhu, Zhenyi Wang, Jiaxin Guo, Kun Ding, Ying Wang, Shen Huang, Xunjie Zhu, Pengjun Xie, Shiming Xiang}
}

\begin{document}

\maketitle

\begin{abstract}
Rubric-based tasks are increasingly addressed through reinforcement learning (RL), with rubric scores used as training rewards.
However, these rewards typically supervise final answers without distinguishing the contributions of intermediate decisions.
Many existing credit assignment methods rely on ground-truth answers to define process rewards, limiting their applicability to open-ended tasks without canonical solutions.
To address this limitation, the proposed rubric-grounded credit uses task requirements as a shared reference for final answer evaluation and process supervision.
The information returned by tools is assessed for the additional support it provides toward satisfying each rubric relative to that rubric's history of accepted support.
By referencing these histories, credit distinguishes new support from evidence already present in the trajectory while recognizing partial support for each rubric.
Dr.Credit uses rubric-grounded credit to supervise intermediate tool turns in an RL framework for deep research agents.
The resulting process advantages are combined with GRPO outcome advantages to guide research decisions while retaining supervision of final-report quality.
Evaluations on four in-domain and out-of-domain benchmarks show that Dr.Credit outperforms the evaluated open deep research baselines on every primary metric and submetric.
Meanwhile, with an 8B-parameter backbone, the trained agent achieves average performance competitive with the evaluated frontier proprietary models.
Further analyses suggest more efficient evidence acquisition and higher-quality reports under limited research-turn budgets, motivating the extension of rubric-grounded process supervision to a broader range of rubric-based tasks.
\end{abstract}

\section{Introduction}

Real-world applications of language models increasingly involve open-ended generation tasks, including novel and screenplay writing~\citep{writingbench}, medical consultation~\citep{healthbench}, and deep research.
Response quality in these settings is difficult to assess through exact answer matching, as evaluation requires nuanced judgments across multiple dimensions.
Task-specific rubrics make these evaluation requirements explicit by specifying the expected content and qualities of a satisfactory response~\citep{rr}.
Rubrics as Rewards (RaR)~\citep{rar} extends this evaluation approach to reinforcement learning (RL) by turning task-specific requirements into training rewards.
In deep research, DR Tulu~\citep{drtulu} and DeepRubric~\citep{deeprubric} train agents with rubric-based rewards, directly optimizing report quality against these requirements.

\begin{figure*}[htb]
    \centering
    \includegraphics[width=\textwidth]{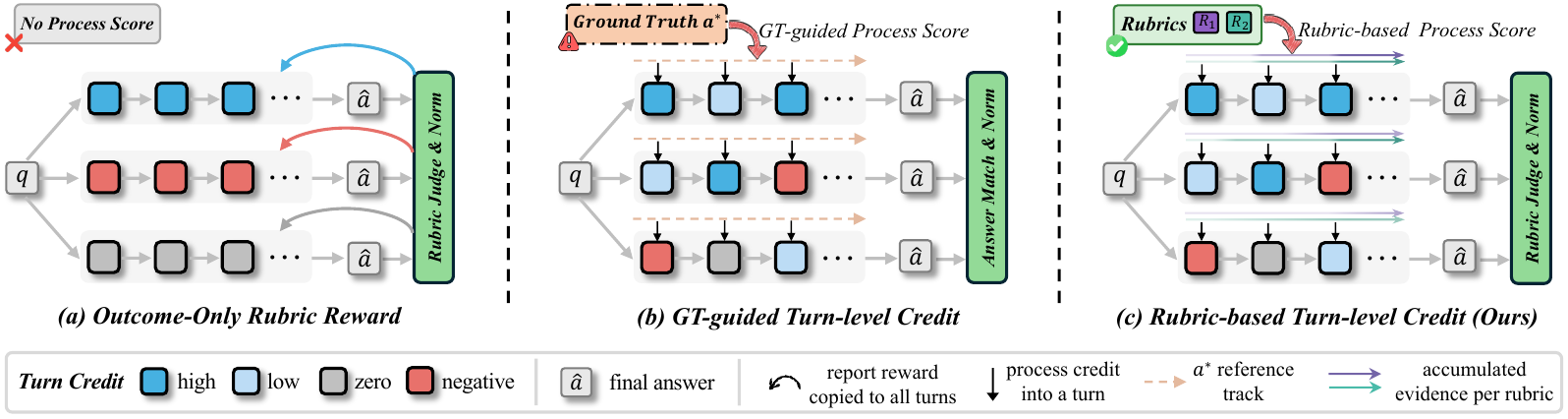}
    \par\vspace{-3mm}
    \caption{\textbf{Credit assignment for rubric-based agent training.}
    (a) Intermediate turns share an outcome advantage.
    (b) Ground-truth answers guide turn-level credit.
    (c) Task rubrics ground turn-level credit in additional evidence support while outcome supervision is retained.
    In (c), each rubric's track darkens as evidence accumulates across turns. Turn colors denote normalized advantages.}
    \label{fig:teaser}
\end{figure*}

Recent work on rubric-based RL has focused on improving rubric quality to provide more reliable supervision for final answers~\citep{drtulu,deeprubric,QUEST}, but leaves credit assignment across intermediate turns unresolved.
When these turns share an outcome advantage, a high-scoring answer can reinforce ineffective decisions, while later synthesis errors can cause useful intermediate steps to be penalized (Figure~\ref{fig:teaser}(a)).
Many existing credit assignment methods use provided ground-truth answers as a reference for assigning intermediate rewards~\citep{IGPO,lapo,abseeker} (Figure~\ref{fig:teaser}(b)).
This dependence limits direct transfer to open-ended tasks where rubrics specify what a satisfactory response should accomplish without prescribing a canonical answer to serve as a reference for turn-level credit assignment.

To address this challenge, the key is to identify a reference for evaluating intermediate turns without relying on a canonical answer.
The rubric set used to evaluate the final answer offers a natural reference, since it defines task-specific requirements that remain applicable across different high-quality responses.
Our core idea is to derive process credit from the additional support that newly obtained information provides toward satisfying each rubric.
We maintain a separate support history for each rubric to assess these contributions and aggregate them into turn-level credit (Figure~\ref{fig:teaser}(c)).
The credits are normalized into process advantages and combined with GRPO outcome advantages, extending rubric-based supervision to intermediate decisions while retaining final-answer evaluation.

This work applies these rubric-grounded credit assignment and optimization mechanisms to deep research, a representative open-ended task with rubric-based evaluation.
The training framework is named Dr.Credit, which uses task-specific rubrics to assign process credit to intermediate tool-use turns.
Turns that extract webpage content are credited for the additional support they provide relative to the existing evidence history for each rubric.
These credits are then attributed to earlier search turns that returned the corresponding URLs.
Search turns can also receive supplementary credit through an independent assessment of rubric support in snippets from results whose webpage content is not subsequently extracted.
The resulting credits enter the preceding policy optimization scheme, providing turn-level supervision alongside the reward for the final report.

To evaluate the effectiveness of Dr.Credit, we conduct extensive experiments on both in-domain and out-of-domain benchmarks with deep research agents.
Results show that Dr.Credit consistently outperforms the evaluated open deep research baselines, suggesting good generalization across benchmarks.
Our main contributions are as follows:
(1) We propose rubric-grounded credit, a general approach to process reward design for agent RL on open-ended tasks with rubric-based evaluation. It extends task rubrics from final-answer evaluation to turn-level credit assignment by assessing additional support relative to per-rubric histories, without requiring a canonical answer.
(2) We apply this approach to deep research and introduce Dr.Credit, a training framework that assigns credit to intermediate research tool turns and integrates the resulting process advantages with GRPO outcome advantages.
(3) With an 8B backbone, Dr.Credit achieves average performance competitive with the evaluated frontier proprietary models. Further analyses indicate that Dr.Credit acquires evidence more efficiently and produces higher-quality reports under limited research-turn budgets.

\section{Preliminaries}
\subsection{Problem formulation}
\label{sec:task_formulation}

Given an open-ended question $q$, a policy $\pi_\theta$ generates a rollout $O=(\tau_0,\ldots,\tau_T)$ through successive rounds of reasoning and interaction with an external environment.
Each tool-interaction turn $\tau_t$ ($t<T$) follows the ReAct paradigm~\citep{react}, comprising \texttt{[think]}, \texttt{[tool call]}, and \texttt{[tool response]}.
Conditioned on the question and the preceding interaction history, the policy generates the reasoning and tool call, while the environment supplies the observation returned by its execution.
The sampled rollout $O_i$ at the bottom of Figure~\ref{fig:main} illustrates how each returned observation is appended to the context, where information gathered so far remains available to subsequent reasoning and tool use.
At the final turn $\tau_T$, \texttt{[think]} is followed by \texttt{[answer]}, yielding a final answer $\hat{a}$ whose quality is then evaluated against the requirements specified by the task's rubric set.

\subsection{Rubric-based reinforcement learning}
\label{sec:rubric_based_rl}

Rubrics as Rewards (RaR)~\citep{rar} turns rubric-based evaluations of final answers into reward signals for training the policy $\pi_\theta$ through reinforcement learning.
For a question $q$, the task-specific rubric set $\mathcal{R}_q$ contains $K$ rubrics, with a nonnegative weight $w_k$ assigned to each rubric and $\sum_{k=1}^{K}w_k>0$.
An LLM judge evaluates the final answer $\hat{a}$ against each rubric, assigning a satisfaction score $s_k(q,\hat{a})\in[0,1]$; the weighted average of these scores provides the rubric reward:
\begin{equation}
\label{eq:rubric_reward}
r_{\mathrm{rubric}}(q,\hat{a})
=\frac{\sum_{k=1}^{K}w_k s_k(q,\hat{a})}{\sum_{k=1}^{K}w_k}.
\end{equation}

Group Relative Policy Optimization (GRPO)~\citep{deepseekmath} converts rollout-level outcome rewards into advantages by normalizing rewards across rollouts sampled for the same question.
Since each advantage is shared by all policy-generated tokens within a rollout, outcome supervision does not explicitly distinguish the contributions of intermediate turns toward satisfying the task's rubrics.

\section{Methodology}
\begin{figure*}[htb]
    \centering
    \includegraphics[width=\textwidth]{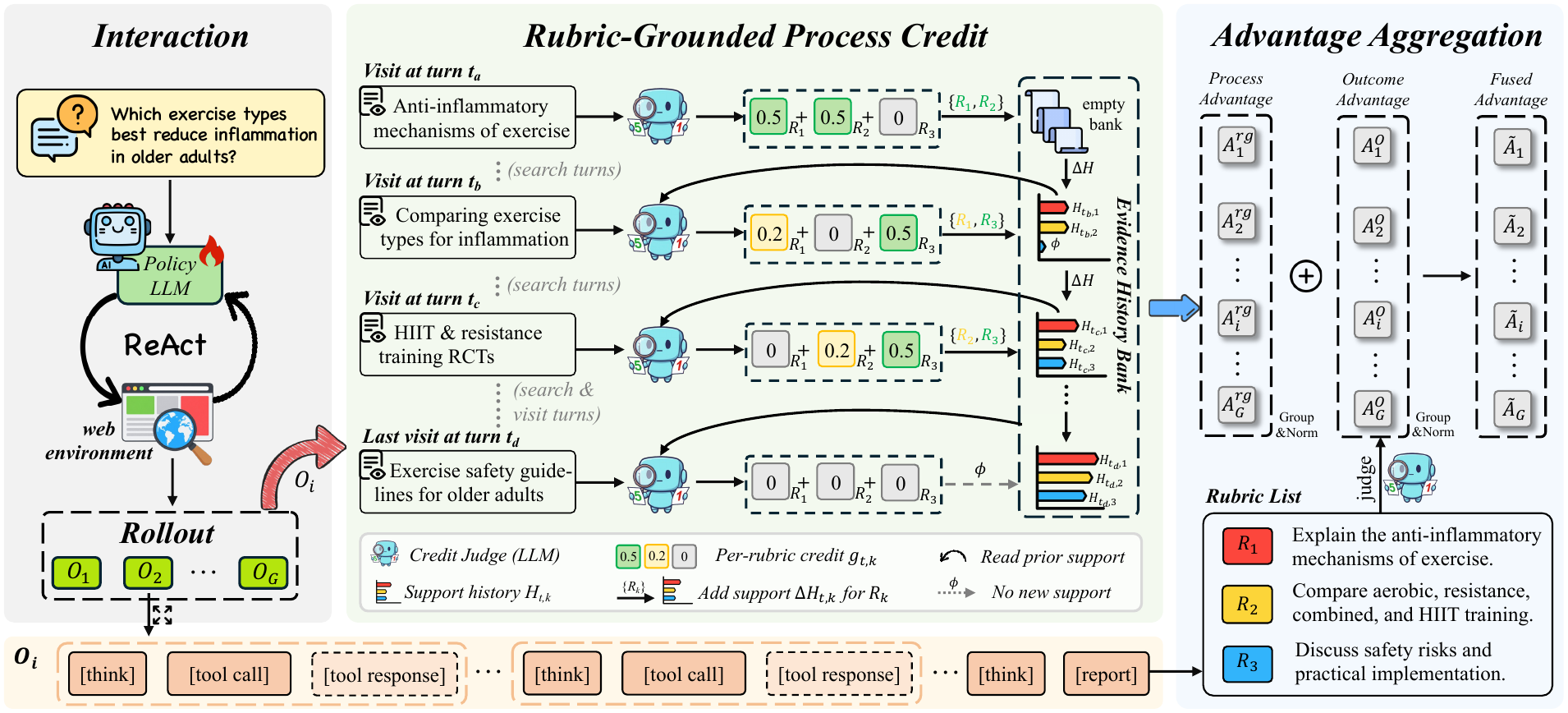}
    \par\vspace{-3mm}
    \caption{\textbf{Overall framework of Dr.Credit}, which incorporates rubric-grounded process credit into reinforcement learning for deep research agents. \textbf{Left:} the policy generates a group of rollouts for each query through ReAct interactions with the web environment. \textbf{Center:} an LLM judge assesses the additional support each tool turn provides relative to per-rubric evidence histories within its rollout, then updates the histories with accepted support points. \textbf{Right:} process credits are normalized across research turns in the rollout group and added to GRPO outcome advantages. \textbf{Bottom:} an example rollout $O_i$. Black dashed boxes mark tool responses masked out of the loss.}
    \label{fig:main}
\end{figure*}

Rubrics can support fine-grained process rewards in reinforcement learning, extending their use beyond evaluating final outputs. However, tool turns cannot be judged by rubric satisfaction in the same way as final answers. Their contributions lie in the information they provide toward satisfying each rubric, but assessing it in isolation can repeatedly credit support established earlier. We therefore ground process credit in the evidence accumulated for each rubric, maintaining support histories against which an LLM judge evaluates additional support from newly obtained information.

Following this principle, we introduce Dr.Credit, a rubric-based credit assignment framework for deep research agents. As illustrated in Figure~\ref{fig:main}, Dr.Credit assigns credit to tool turns based on their contributions toward satisfying the task's rubrics and combines the resulting process advantages with GRPO outcome advantages for policy optimization. We first formalize the underlying rubric-grounded process credit and the maintenance of support histories (Section~\ref{sec:credit_formulation}), then describe how the resulting credits enter policy optimization (Section~\ref{sec:policy_optimization}). Section~\ref{sec:deepresearch_instantiation} presents Dr.Credit as a concrete instantiation of these mechanisms for a basic deep research agent.

\subsection{Rubric-Grounded Process Credit}
\label{sec:credit_formulation}

\noindent\textbf{Rubric-Grounded Evidence Support.}
For a question $q$ with rubric set $\mathcal R_q$, let $e_t$ denote the information in the environment's response at tool turn $t$ of rollout $O$. For each rubric, let $\mathcal S_k(q)$ be the space of evidence-grounded support statements for the $k$-th rubric, including facts, premises, relations, and qualifications that help fulfill its requirement without prescribing a canonical answer or an action sequence. While $s_k(q,\hat a)$ measures how well the final answer satisfies the rubric, process assessment identifies the support that $e_t$ establishes toward meeting the same requirement.

\noindent\textbf{Per-Rubric Support Histories.}
To distinguish new support from previously established information, we maintain a history $H_{t,k}\subseteq\mathcal S_k(q)$ containing the support points accepted for the $k$-th rubric before turn $t$. These histories start empty, $H_{0,k}=\varnothing$, and evolve independently within each rollout. Their collection, $\mathcal H_t=(H_{t,1},\ldots,H_{t,K})$, forms the history bank in Figure~\ref{fig:main}, preserving prior support as the rubric-specific reference for assessing subsequent information.

\noindent\textbf{History-Aware Credit Assignment.}
The assessor $\mathcal J$ compares the current information $e_t$ with histories $\mathcal H_t$, evaluating additional support per rubric and identifying the support increments:
\begin{equation}
\label{eq:rubric_assessment}
(\mathbf g_t,\Delta\mathcal H_t)
=\mathcal J(q,\mathcal R_q,e_t,\mathcal H_t).
\end{equation}
Here, $\mathbf g_t=(g_{t,k})_{k=1}^{K}$ contains nonnegative per-rubric contributions, and $\Delta\mathcal H_t=(\Delta H_{t,k})_{k=1}^{K}$ contains the corresponding support increments, with $\Delta H_{t,k}\subseteq\mathcal S_k(q)$. Each $g_{t,k}$ measures the support that $e_t$ adds relative to $H_{t,k}$, with zero assigned when no additional support is established; the sum of these contributions defines the rubric-grounded credit for the current turn as
\begin{equation}
\label{eq:general_process_credit}
c_t^{rg}=\sum_{k=1}^{K}g_{t,k}.
\end{equation}

Once the contributions have been assessed against the existing histories, the accepted support increments update the reference available to subsequent turns through the per-rubric union
\begin{equation}
\label{eq:visit_history_update}
H_{t+1,k}=H_{t,k}\cup\Delta H_{t,k}.
\end{equation}
Assessment reads the history before the update, so the current evidence is compared with prior support before its accepted points enter the history; an empty increment leaves that history unchanged.

\subsection{Policy Optimization with Process Credit}
\label{sec:policy_optimization}

Process credits measure additional rubric support from each turn, while the outcome reward evaluates the final answer. To use both signals in policy optimization, the old policy $\pi_{\mathrm{old}}$ samples $G>1$ rollouts for each question--rubric pair $(q,\mathcal R_q)$ in the training set $\mathcal D$. Let $c_{i,t}$ be the process credit for non-final turn $t<T_i$ of rollout $O_i$, where $T_i$ indexes its final turn, and let $r_i$ denote its rollout-level outcome reward. We normalize both signals separately within each rollout group to obtain
\begin{equation}
\label{eq:process_advantage}
A_{i,t}^{rg}=\frac{c_{i,t}-\mu^{rg}}{\sigma^{rg}},
\qquad
A_i^{O}=\frac{r_i-\mu^{O}}{\sigma^{O}}.
\end{equation}
The process statistics $\mu^{rg}$ and $\sigma^{rg}$ are the mean and population standard deviation over all non-final policy-turn credits in the rollout group, including zero-credit turns, with equal weight per turn regardless of the length of its rollout. The outcome statistics $\mu^O$ and $\sigma^O$ retain the GRPO mean and sample standard deviation over the $G$ rollout rewards obtained for the same question.

These separately normalized advantages provide the two inputs to the fusion step on the right of Figure~\ref{fig:main}, where the process term supplements the outcome advantage on each non-final turn:
\begin{equation}
\label{eq:fused_advantage}
\widetilde A_{i,t}=
\begin{cases}
A_i^{O}+A_{i,t}^{rg}, & 0\leq t<T_i,\\
A_i^{O}, & t=T_i.
\end{cases}
\end{equation}

To apply this turn-level signal to token-level updates, let $\mathcal P_i$ contain the policy-generated token positions in $O_i$, and let $t(i,j)$ identify the turn containing token $x_{i,j}$. All tokens in that turn share $\widetilde A_{i,t(i,j)}$, while the policy ratio remains $\rho_{i,j}(\theta)=\frac{\pi_\theta(x_{i,j}\mid q,x_{i,<j})}{\pi_{\mathrm{old}}(x_{i,j}\mid q,x_{i,<j})}$, where $x_{i,<j}$ includes earlier policy tokens and environment observations. Substituting $\widetilde A_{i,t(i,j)}$ into the GRPO surrogate gives the following objective, averaged over all policy-generated tokens in the sampled rollout group
\begin{equation}
\label{eq:drcredit_objective}
\begin{aligned}
\mathcal{J}_{\mathrm{Dr.Credit}}(\theta)
&=\mathbb{E}_{(q,\mathcal{R}_q)\sim\mathcal{D},\,\{O_i\}\sim\pi_{\mathrm{old}}(\cdot\mid q)}
\Bigg[\frac{1}{\sum_{i=1}^{G}|\mathcal{P}_i|}
\sum_{i=1}^{G}\sum_{j\in\mathcal{P}_i}
\min\!\Big(
\rho_{i,j}(\theta)\widetilde A_{i,t(i,j)},\\
&\qquad
\operatorname{clip}\!\big(\rho_{i,j}(\theta),1-\epsilon,1+\epsilon\big)
\widetilde A_{i,t(i,j)}\Big)
-\beta\mathbb{D}_{\mathrm{KL}}(\pi_\theta\Vert\pi_{\mathrm{ref}})
\Bigg].
\end{aligned}
\end{equation}
Here, $\epsilon$ controls probability-ratio clipping, and the KL term denotes token-averaged regularization against reference policy $\pi_{\mathrm{ref}}$, with coefficient $\beta$. The dashed masks in Figure~\ref{fig:main} exclude environment observations from the loss while retaining them in the policy's conditioning context.

\subsection{Rubric-Grounded Credit for Deep Research}
\label{sec:deepresearch_instantiation}
\begin{figure*}[ht]
    \centering
    \includegraphics[width=\textwidth]{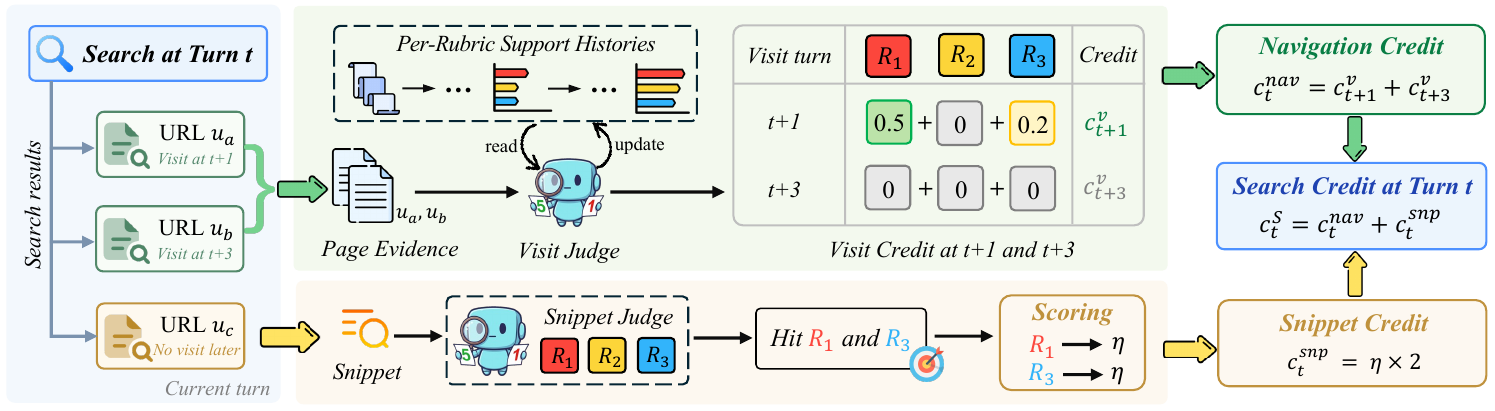}
    \par\vspace{-3mm}
    \caption{\textbf{Composite credit for \texttt{Search} turns in Dr.Credit.} \textbf{Upper green branch:} history-aware Visit credits are attributed to the search through URL matching. \textbf{Lower yellow branch:} an independent judge matches snippets from results without later visits to rubrics, without reading or updating Visit histories. Navigation and snippet credits are added to obtain Search credit.}
    \label{fig:credit}
\end{figure*}

Dr.Credit instantiates the preceding credit assignment and policy optimization mechanisms for a basic deep research agent equipped with only two common tools. \texttt{Search} discovers candidate sources and returns snippets, while \texttt{Visit} extracts the content of selected webpages. \texttt{Visit} provides the main body of evidence for the final report and is the primary focus of our credit assessment. \texttt{Search} receives credit for discovering sources that yield additional support and for evidence available directly in its snippets, as illustrated in Figure~\ref{fig:credit}. The resulting Visit and Search credits, denoted $c_t^v$ and $c_t^s$, provide $c_{i,t}$ for policy optimization in Section~\ref{sec:policy_optimization}.

\phantomsection\label{sec:visit_credit}
For a \texttt{Visit} turn, the page evidence serves as $e_t$. An LLM judge implements $\mathcal J$ by assessing the additional support this evidence provides relative to each rubric's history $H_{t,k}$ and assigning a contribution score $g_{t,k}$. These per-rubric scores are aggregated into Visit credit $c_t^v$ following the summation in Equation~\ref{eq:general_process_credit}. For positive assessments, the judge extracts support points in the current page to form $\Delta H_{t,k}$, which updates the corresponding history after scoring according to Equation~\ref{eq:visit_history_update}.

\phantomsection\label{sec:search_credit}
For a \texttt{Search} turn, we match normalized result URLs to later visits within the same rollout and aggregate the matched Visit credits into navigation credit $c_t^{\mathrm{nav}}$. The upper branch of Figure~\ref{fig:credit} illustrates this attribution for visits to $u_a$ and $u_b$. Results without later visits enter the complementary snippet branch, where an independent LLM judge identifies rubric support directly in their snippets. Each accepted rubric match contributes $\eta$ to snippet credit $c_t^{\mathrm{snp}}$; in the illustrated example, the snippet from $u_c$ supports $R_1$ and $R_3$, yielding two such contributions. The brief, scattered information in snippets motivates coarse rubric-level matching without reading or updating Visit histories. Navigation and snippet credits are combined additively into Search credit $c_t^s$.

\section{Experiments}
\begin{table*}[ht]
\centering
\vspace{-3.5mm}
\caption{Overall performance on four deep research benchmarks.
Dr.Credit outperforms all open deep research baselines on every metric, including all submetrics, and is competitive with proprietary systems.
Average is the unweighted mean of the four benchmark-level scores.
Bold denotes the best scores across the lower three model groups.
$^{*}$ and $^{\dagger}$ indicate results taken from DeepRubric~\citep{deeprubric} and ResearchRubrics~\citep{rr}, respectively.}
\vspace{1mm}
\label{tab:main_results}
\begingroup
\small
\setlength{\tabcolsep}{4pt}
\renewcommand{\arraystretch}{1.12}
\resizebox{\textwidth}{!}{%
\begin{tabular}{@{}l*{11}{c}@{}}
\toprule
\multirow{2}{*}{\textbf{Methods}}
& \multicolumn{3}{c}{\textbf{DeepRubric (val)}}
& \multirow{2}{*}{\textbf{ResearchQA}}
& \multicolumn{5}{c}{\textbf{DeepResearchBench}}
& \multirow{2}{*}{\textbf{ResearchRubrics}}
& \multirow{2}{*}{\textbf{Average}} \\
\cmidrule(lr){2-4}
\cmidrule(lr){6-10}
& \textbf{Overall} & \textbf{Factual} & \textbf{Logical}
&
& \textbf{Overall} & \textbf{Comp.} & \textbf{Depth}
& \textbf{Instr.} & \textbf{Read.}
& & \\
\midrule
\multicolumn{12}{@{}l}{\textbf{\textit{Frontier Proprietary Models}}} \\
\addlinespace[2pt]
\rowcolor[gray]{0.94}
Kimi K3 + Our Tools
& 71.5 & 55.5 & 90.2 & 76.7
& 48.4 & 47.2 & 47.5 & 50.2 & 49.5
& 58.0 & 63.6 \\
\rowcolor[gray]{0.94}
GPT-5.6-luna + Our Tools
& 73.2 & 55.6 & 93.9 & 73.2
& 48.8 & 47.3 & 48.4 & 50.4 & 49.8
& 56.6 & 63.0 \\
\rowcolor[gray]{0.94}
DeepSeek-V4-Pro + Our Tools
& 68.4 & 52.7 & 87.0 & 78.0
& 44.5 & 43.5 & 43.7 & 46.1 & 45.6
& 50.8 & 60.4 \\
\rowcolor[gray]{0.94}
Opus 4.8 + Our Tools
& 71.0 & 55.2 & 89.7 & 72.9
& 45.5 & 43.6 & 44.8 & 47.7 & 47.6
& 51.5 & 60.2 \\
\rowcolor[gray]{0.94}
GLM-5.3 + Our Tools
& 68.9 & 54.2 & 85.9 & 72.1
& 45.6 & 44.4 & 45.5 & 46.1 & 46.0
& 52.5 & 59.8 \\
\rowcolor[gray]{0.94}
Perplexity Deep Research
& -- & -- & -- & $75.3^{*}$
& $42.3^{*}$ & $40.7^{*}$ & $39.3^{*}$ & $46.4^{*}$ & $44.3^{*}$
& $48.7^{\dagger}$ & -- \\
\rowcolor[gray]{0.94}
Gemini Deep Research
& -- & -- & -- & $68.5^{*}$
& $48.8^{*}$ & $48.5^{*}$ & $48.5^{*}$ & $49.2^{*}$ & $49.4^{*}$
& $61.5^{\dagger}$ & -- \\
\rowcolor[gray]{0.94}
OpenAI Deep Research
& -- & -- & -- & $79.2^{*}$
& $46.9^{*}$ & $46.8^{*}$ & $45.2^{*}$ & $49.2^{*}$ & $47.1^{*}$
& $59.7^{\dagger}$ & -- \\
\midrule
\multicolumn{12}{@{}l}{\textbf{\textit{Naive RAG}}} \\
\addlinespace[2pt]
Qwen3-8B + RAG
& 31.6 & 20.9 & 44.2 & 46.0
& 22.3 & 19.0 & 14.3 & 32.8 & 26.6
& 26.2 & 31.5 \\
Qwen3.5-9B + RAG
& 42.5 & 25.4 & 63.4 & 47.2
& 26.1 & 22.5 & 18.6 & 36.1 & 31.3
& 32.9 & 37.2 \\
Qwen3.6-35B-A3B + RAG
& 57.7 & 38.1 & 81.0 & 59.1
& 36.7 & 33.7 & 32.3 & 44.3 & 39.3
& 43.6 & 49.3 \\
\midrule
\multicolumn{12}{@{}l}{\textbf{\textit{Open Deep Research Models}}} \\
\addlinespace[2pt]
ASearcher-Web-7B
& 12.3 & 7.8 & 16.9 & $19.4^{*}$
& $7.8^{*}$ & $5.1^{*}$ & $1.7^{*}$ & $15.2^{*}$ & $11.8^{*}$
& 7.6 & 11.8 \\
Search-R1-7B
& 10.3 & 7.2 & 14.3 & $27.9^{*}$
& $9.5^{*}$ & $5.2^{*}$ & $2.1^{*}$ & $18.6^{*}$ & $16.8^{*}$
& 5.2 & 13.2 \\
WebExplorer-8B
& 45.0 & 37.0 & 53.3 & $64.8^{*}$
& $36.7^{*}$ & $33.7^{*}$ & $28.5^{*}$ & $45.7^{*}$ & $42.2^{*}$
& 29.5 & 44.0 \\
Tongyi DeepResearch-30B-A3B
& 57.3 & 44.3 & 72.4 & $66.7^{*}$
& $40.6^{*}$ & $39.1^{*}$ & $34.3^{*}$ & $46.8^{*}$ & $45.4^{*}$
& 37.1 & 50.4 \\
\midrule
\multicolumn{12}{@{}l}{\textbf{\textit{Qwen3-8B-Based Deep Research Agents}}} \\
\addlinespace[2pt]
Qwen3-8B + Our Tools
& 40.5 & 30.0 & 52.8 & 48.1
& 28.8 & 26.6 & 23.3 & 35.4 & 33.0
& 33.0 & 37.6 \\
DeepRubric-8B + Our Tools
& 62.5 & 43.6 & 85.6 & 73.5
& 41.6 & 39.7 & 39.1 & 45.4 & 44.3
& 43.1 & 55.2 \\
DR Tulu-8B + Our Tools
& 60.4 & 41.7 & 83.3 & 72.9
& 42.9 & 41.2 & 42.4 & 45.6 & 43.2
& 45.8 & 55.5 \\
Qwen3-8B-SFT
& 59.3 & 45.2 & 75.3 & 69.2
& 40.3 & 39.3 & 34.6 & 45.4 & 41.8
& 40.9 & 52.4 \\
Qwen3-8B-GRPO
& 67.4 & 52.5 & 84.0 & 75.4
& 43.2 & 42.1 & 40.9 & 46.8 & 44.0
& 46.8 & 58.2 \\
\rowcolor[rgb]{0.89,0.94,0.98}
\textbf{Dr.Credit (Ours)}
& \textbf{70.6} & \textbf{54.0} & \textbf{90.0} & \textbf{79.8}
& \textbf{46.1} & \textbf{45.1} & \textbf{44.7} & \textbf{48.7} & \textbf{46.7}
& \textbf{50.3} & \textbf{61.7} \\
\bottomrule
\end{tabular}%
}
\endgroup
\par\vspace{-3mm}
\end{table*}

\subsection{Experimental Setup}
\label{sec:experimental_setup}

\noindent\textbf{Datasets \& Metrics.}
DeepRubric~\cite{deeprubric} provides our training data, pairing research queries with evidence-grounded rubric sets; we randomly hold out 128 examples for in-domain validation.
Out-of-domain evaluation covers three research benchmarks: ResearchQA~\cite{researchqa} for scholarly question answering, DeepResearchBench~\cite{drb} for comprehensive report generation, and ResearchRubrics for open-ended tasks with expert-written rubrics.
We report weighted rubric satisfaction on the DeepRubric validation set, rubric coverage on ResearchQA, weighted rubric compliance on ResearchRubrics, and report-quality scores on DeepResearchBench, all on a 100-point scale.
Data construction and evaluation settings are detailed in Appendix~\ref{app:setup}.

\noindent\textbf{Baselines.}
Our baselines cover four groups: (1) frontier models equipped with our tools, including Kimi K3~\cite{kimi_k3}, DeepSeek-V4-Pro~\cite{deepseek_v4}, and GLM-5.3~\cite{glm5}, together with commercial deep research services; (2) Qwen-based naive RAG~\cite{qwen3}; (3) open search agents, including ASearcher~\cite{asearcher}, Search-R1~\cite{searchr1}, WebExplorer~\cite{webexplorer}, and Tongyi DeepResearch~\cite{tongyi_deepresearch}; and (4) open deep research agents built on Qwen3-8B, including DeepRubric-8B~\cite{deeprubric}, DR Tulu-8B~\cite{drtulu}, and our SFT and GRPO baselines, with the base Qwen3-8B agent included as a reference.
The GRPO baseline uses outcome rewards without process credit and shares the same SFT initialization, training data, outcome reward, training budget, and evaluation setup as Dr.Credit.
Baseline evaluation settings and result provenance are detailed in Appendix~\ref{app:inference}.

\noindent\textbf{Implementation Details.}
For credit assignment (Section~\ref{sec:deepresearch_instantiation}), \texttt{Visit} assessments of no additional support, partial new support, and high-value new evidence correspond to $g_{t,k}=0$, $0.2$, and $0.5$, respectively, while each accepted rubric match in a \texttt{Search} snippet contributes $\eta=0.1$ to $c_t^{\mathrm{snp}}$.
To keep turn-level credits on a shared scale before process-advantage normalization, the sums defining Visit credit $c_t^v$, navigation credit $c_t^{\mathrm{nav}}$, and Search credit $c_t^s$ are capped at $1$.
An offline analysis in Appendix~\ref{app:encodings} reports similar normalized process advantages across alternative parameter settings.

With these credit settings, RL starts from a Qwen3-8B checkpoint obtained after two epochs of SFT on 1,505 filtered research trajectories.
Dr.Credit is then trained on 7,149 query--rubric pairs for 200 steps, with 32 queries per batch and $G=8$ rollouts per query.
For both process and outcome supervision, Qwen3.6-35B-A3B performs history-aware \texttt{Visit} assessment and independent \texttt{Search} snippet matching, as well as evaluating rubric satisfaction and citation quality in final reports.
The corresponding judge prompts are provided in Appendix~\ref{app:prompts}, while Appendix~\ref{app:setup} documents data construction, training and interaction settings, and the definitions of each outcome reward component.

\subsection{Overall Performance}
\label{sec:overall_performance}

Table~\ref{tab:main_results} shows that Dr.Credit is competitive with frontier models, surpassing DeepSeek-V4-Pro, Opus 4.8, and GLM-5.3 in average score while approaching Kimi K3 and GPT-5.6-luna.
With an 8B backbone, Dr.Credit achieves an average score of 61.7 and outperforms all models in the Naive RAG and Open Deep Research Models groups across every metric and submetric, including larger models such as Qwen3.6-35B-A3B and Tongyi DeepResearch-30B-A3B.

Among deep research agents built on the same Qwen3-8B base model, Dr.Credit achieves the highest scores on all four benchmarks and their submetrics.
The results for DeepRubric-8B and DR Tulu-8B in Table~\ref{tab:main_results} are from our evaluations using the same web search and web visit tools as Dr.Credit (see Appendix~\ref{app:inference}).
Under matched training and evaluation settings, Dr.Credit improves over the GRPO baseline on all four benchmarks, with a 3.5-point average gain from adding process credit.
Beyond the in-domain DeepRubric validation set, Dr.Credit achieves clear gains on all three external benchmarks, suggesting that our credit assignment framework improves out-of-domain generalization.

\subsection{Ablation Study}
\label{sec:ablation_study}

To understand which design choices contribute to Dr.Credit's performance, we conduct three groups of ablation experiments. Specifically, we evaluate the process credit design, the benefits of combining outcome and process supervision, and the contribution of per-rubric evidence history.

\begin{table}[htb]
\centering
\vspace{-4mm}
\caption{\textbf{Ablation of process credit designs.}
Visit denotes history-aware credit for Visit turns; gray shading marks the final Dr.Credit configuration.
DRub, RQA, DRB, and RR denote DeepRubric (val), ResearchQA, DeepResearchBench, and ResearchRubrics, respectively.
Avg is the arithmetic mean of the four benchmark scores; \textbf{bold} marks the best score per benchmark and the highest Avg.}
\label{tab:ablation_components}
\vspace{1.5mm}
\begingroup
\footnotesize
\setlength{\tabcolsep}{2.5pt}
\renewcommand{\arraystretch}{1.13}
\newcommand{\ablationComponentTable}{%
\begin{tabular}{*{4}{>{\centering\arraybackslash}p{0.10\linewidth}}|*{5}{>{\hspace{2.5pt}}r<{\hspace{2.5pt}}}}
\toprule
Visit & Navigation & Snippets & Discount & DRub & RQA & DRB & RR & Avg \\
\midrule
\ding{51} & \ding{55} & \ding{55} & \ding{55} & 68.8 & 76.8 & 44.7 & 48.6 & 59.7 \\
\ding{51} & \ding{51} & \ding{55} & \ding{55} & \textbf{70.8} & 78.4 & 45.0 & 47.2 & 60.4 \\
\rowcolor[gray]{0.93}
\ding{51} & \ding{51} & \ding{51} & \ding{55} & 70.6 & \textbf{79.8} & \textbf{46.1} & \textbf{50.3} & \textbf{61.7} \\
\ding{51} & \ding{51} & \ding{55} & \ding{51} & 68.2 & 77.3 & 44.5 & 45.9 & 59.0 \\
\ding{51} & \ding{51} & \ding{51} & \ding{51} & 68.0 & 78.0 & 45.3 & 48.1 & 59.9 \\
\bottomrule
\end{tabular}%
}
\ablationComponentTable
\endgroup
\par\vspace{-2mm}
\end{table}

\noindent\textbf{Process Credit Design.}
Table~\ref{tab:ablation_components} reports the performance of agents trained with different process credit designs. 
Relative to Visit-only credit, adding navigation attribution from subsequent Visits improves scores on DeepRubric, ResearchQA, and DRB, while reducing the score on RR.
Adding independent snippet scoring for unvisited search results further improves scores on ResearchQA, DRB, and RR, yielding the highest four-benchmark average of 61.7, with only a 0.2-point decrease on DeepRubric.
These comparisons support assigning credit to both source discovery and evidence available directly in search snippets.
Discounted propagation with $\gamma=0.95$ lowers scores on all four benchmarks, both with and without snippet scoring, relative to the corresponding variants without propagation.
Together, these results support the final process credit design adopted in Dr.Credit.

\begingroup
\setlength{\intextsep}{0pt}
\begin{wraptable}{r}{0.52\textwidth}
\setlength{\abovecaptionskip}{0pt}
\centering
\caption{Ablations of Dr.Credit:
(a) removing either outcome or process advantage from research-turn supervision;
(b) removing evidence history when computing process credit.
Each ablation is relative to full Dr.Credit.
Values are score changes, with $\Delta$Avg$_3$ averaged across the three benchmarks.}
\label{tab:ablation_advantage_history}
\small
\setlength{\tabcolsep}{1.8pt}
\renewcommand{\arraystretch}{1.13}
\begin{tabular*}{\linewidth}{@{\extracolsep{\fill}}lrrrr@{}}
\toprule
Removed & $\Delta$DRub & $\Delta$RQA & $\Delta$DRB & $\Delta$Avg$_3$ \\
\midrule
\multicolumn{5}{@{}l}{\textit{(a) Outcome and Process Supervision}} \\
\addlinespace[2pt]
Process $A^{rg}$ & -3.2 & -4.4 & -2.9 & -3.5 \\
Outcome $A^{O}$ & -1.2 & -3.6 & -2.1 & -2.3 \\
\midrule
\multicolumn{5}{@{}l}{\textit{(b) Evidence History}} \\
\addlinespace[2pt]
History & -3.0 & -3.9 & -2.1 & -3.0 \\
\bottomrule
\end{tabular*}
\end{wraptable}
\noindent\textbf{Outcome and Process Supervision.}
As shown in Table~\ref{tab:ablation_advantage_history}(a), removing the process advantage $A^{rg}$ reduces Dr.Credit to GRPO and lowers the three-benchmark average by 3.5 points.
This gap demonstrates the value of rubric-grounded process credit as additional supervision for research turns.
However, relying on process supervision alone for these turns also lowers the average by 2.3 points.
These results suggest complementary roles for the two signals: process credit guides evidence acquisition at individual turns, while outcome supervision helps align these decisions with the overall quality of final reports.

\noindent\textbf{Effect of Evidence History.}
Table~\ref{tab:ablation_advantage_history}(b) shows that computing process credit without evidence history lowers scores on all three benchmarks, with an average decrease of 3.0 points.
A page can support a rubric without extending the evidence already available, so scoring it in isolation risks rewarding repeated retrieval.
The paired scoring analysis in Appendix~\ref{app:history} compares Visit credits and rubric-level judgments with and without prior support on fixed trajectories.
These diagnostics complement the training ablation, suggesting that per-rubric histories help supervise the additional evidence acquired at each turn while accounting for support already available within the trajectory.
\par
\endgroup

\subsection{In-Depth Analysis}
\label{sec:efficiency_analysis}

\begin{figure}[htb]
    \centering
    \includegraphics[width=\linewidth]{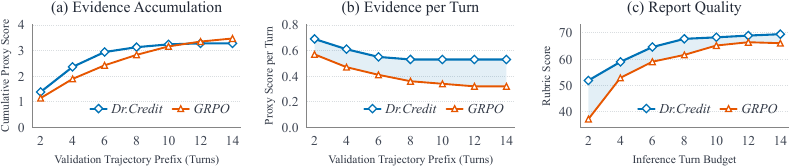}
    \par\vspace{-3mm}
    \caption{Evidence acquisition and report quality on DeepRubric.
    (a) Mean cumulative evidence proxy score.
    (b) Mean proxy score per actual research turn.
    (c) Report rubric scores under inference-time turn budgets.
    Blue shading in (b) and (c) indicates differences between method means.}
    \label{fig:efficiency_analysis}
\end{figure}

To investigate what underlies Dr.Credit's strong performance, we analyze how it acquires evidence along research trajectories and assess report quality under constrained turn budgets.
Figure~\ref{fig:efficiency_analysis} summarizes these analyses on the DeepRubric validation set.
Evidence acquisition is tracked using a credit-based proxy that accumulates uncapped credit from retrieved pages and search snippets, excluding navigation credit.
For each trajectory prefix, the proxy includes only contributions from content returned up to that point.
Scoring and aggregation details are provided in Appendix~\ref{app:evidence_protocol}.

Under this proxy, Dr.Credit accumulates evidence more rapidly in early research turns, reaching a cumulative score of 2.94 within 6 turns compared with GRPO's 2.83 within 8 (Figure~\ref{fig:efficiency_analysis}(a)).
Dr.Credit leads at every measured prefix through ten turns, while GRPO slightly surpasses it at twelve and fourteen turns.
Together with the decline in estimated research-turn counts during training (Appendix~\ref{app:training_behavior}, Figure~\ref{fig:appendix_training_diagnostics}(b)), these results suggest that Dr.Credit maintains effective evidence acquisition as its trajectories become shorter.
Figure~\ref{fig:efficiency_analysis}(b) reports mean per-turn proxy scores by dividing each trajectory's cumulative score by the number of research turns actually completed within the prefix and then averaging these ratios across trajectories (see Appendix~\ref{app:evidence_protocol} for detailed calculations).
Dr.Credit scores higher at every measured prefix, reaching 0.53 versus GRPO's 0.36 at eight turns (+47.2\%), indicating more efficient evidence acquisition per research turn under this proxy.

Separate evaluations under inference-time turn budgets assess final-report quality (Figure~\ref{fig:efficiency_analysis}(c)), following the report-generation protocol in Appendix~\ref{app:evidence_protocol}.
Dr.Credit outperforms GRPO at all seven tested budgets.
With an eight-turn budget, it achieves a rubric score of 67.69, exceeding GRPO's best observed score of 66.41 at twelve turns.
Dr.Credit thus exceeds GRPO's best observed quality with a one-third smaller allowed turn budget.
These results suggest that Dr.Credit follows more efficient research trajectories to produce higher-quality reports under constrained turn budgets.

\subsection{Computational Efficiency}
\label{sec:computational_efficiency}
\begin{figure}[htb]
    \centering
    \includegraphics[width=0.85\linewidth]{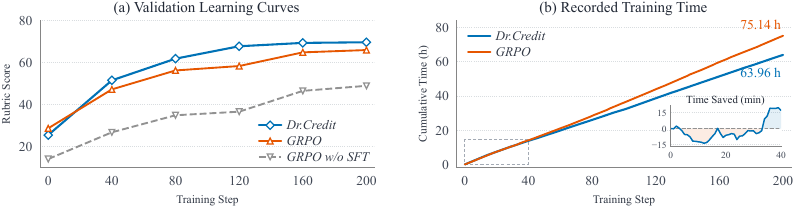}
    \par\vspace{-3mm}
    \caption{Learning progress and recorded training time.
    (a) Validation rubric scores on DeepRubric; GRPO without SFT is included as an initialization reference.
    (b) Cumulative recorded training-step time over 200 steps, including rollout collection, reward scoring, and policy updates.
    The inset shows GRPO time minus Dr.Credit time over the first 40 steps, in minutes.}
    \par\vspace{-2.5mm}
    \label{fig:learning_training_efficiency}
\end{figure}

Although Dr.Credit achieves strong performance, its process-credit scoring may introduce substantial computational overhead during training.
Figure~\ref{fig:learning_training_efficiency} therefore compares the learning progress and cumulative training time of Dr.Credit and GRPO over 200 steps, including process-credit scoring.

\noindent\textbf{Learning progress.}
As illustrated in Figure~\ref{fig:learning_training_efficiency}(a), our rubric-grounded credit assignment enables faster empirical convergence and higher final validation performance than GRPO baseline.
Dr.Credit surpasses GRPO's final validation score by step 120 and finishes at step 200 with a 3.2-point advantage.
The supplementary training curves in Appendix~\ref{app:training_behavior} show consistent improvements.

\noindent\textbf{Training time.}
Figure~\ref{fig:learning_training_efficiency}(b) shows an early cumulative-time disadvantage for Dr.Credit.
This is expected given the additional workload of process-credit scoring, as reported in Table~\ref{tab:process_credit_cost}.
From step 34 onward, however, Dr.Credit maintains a lower cumulative training time than GRPO, saving 11.18 hours (14.9\%) over 200 steps with scoring included.
Dr.Credit learns more efficient research behavior that reduces interaction and evaluation work, while asynchronous execution in the training framework allows scoring to overlap with rollout generation.
These factors help explain the time savings under the measured setup, with a detailed analysis provided in Appendix~\ref{app:training_cost}.

\section{Related Work}

\noindent\textbf{Rubric-Based Rewards for Deep Research.}
Instance-specific rubrics and checklists provide evaluation criteria for open-ended tasks, with recent work improving their coverage and discriminability~\cite{rr, viswanathan2026checklists, shen2026rethinking}.
In deep research, DR Tulu~\cite{drtulu} evolves rubrics using retrieved evidence and on-policy rollouts, while DeepRubric~\cite{deeprubric} and DR-Rubric~\cite{mei2026deep} ground rubrics in evidence from structured expansion or agentic search.
Learning Query-Specific Rubrics~\cite{lv2026learning} aligns rubric generation with human preferences for research reports, and QUEST~\cite{QUEST} uses rubric trees to define rewards across research tasks.
RubricEM~\cite{rubricem} extends rubric supervision to intermediate research stages, assigning a shared advantage to all policy tokens within each stage.

\noindent\textbf{Fine-Grained Credit Assignment.}
Classical temporal credit assignment uses TD($\lambda$)~\cite{sutton1988td} to propagate prediction errors through eligibility traces; generalized advantage estimation~\cite{schulman2016gae} builds on this principle to balance bias and variance in policy-gradient estimation.
RUDDER~\cite{arjona2019rudder} addresses delayed rewards by redistributing returns according to changes in predicted return.
For LLM agents, GiGPO~\cite{GIGPO} compares actions from repeated anchor states across trajectories; HCAPO~\cite{HCAPO} uses hindsight reasoning to refine step-level value estimates; and HiPER~\cite{Hiper} estimates advantages for both planning and execution.
A complementary line measures progress toward a target answer: IGPO~\cite{IGPO} rewards increases in the policy's probability of generating the correct answer; TIPS~\mbox{\cite{TIPS}} shapes rewards using a teacher model's answer potential; and $\Delta$Belief-RL~\cite{Belief-RL} rewards changes in the agent's belief in the target solution.

\section{Conclusion}
\label{sec:conclusion}

In this work, we propose rubric-grounded credit, which extends task rubrics from final-answer evaluation to process supervision. Intermediate steps receive credit for the additional support they provide relative to each rubric's evidence history, without requiring a canonical answer. Applied to deep research, this approach underlies Dr.Credit, a training framework combining process and GRPO outcome advantages to supervise research decisions and final-report quality. Evaluations across four benchmarks show that Dr.Credit outperforms the evaluated open deep research baselines and generalizes well to out-of-domain tasks. Even with only 8B parameters, the trained Qwen3-8B agent achieves average performance competitive with that of the evaluated frontier proprietary models. Beyond these benchmark results, further analyses suggest that the trained agent acquires useful evidence more efficiently and produces higher-quality reports under the same research-turn budgets. These findings support rubric-based supervision of evidence acquisition and motivate extending rubric-grounded credit to other open-ended tasks trained with rubric-based reinforcement learning.

\bibliography{iclr2027_conference}
\bibliographystyle{iclr2027_conference}

\clearpage
\appendix
\setlength{\parskip}{4pt}
\setcounter{topnumber}{3}
\setcounter{bottomnumber}{2}
\setcounter{totalnumber}{4}
\renewcommand{\topfraction}{0.95}
\renewcommand{\bottomfraction}{0.95}
\renewcommand{\textfraction}{0.04}
\renewcommand{\floatpagefraction}{0.80}
\setlength{\textfloatsep}{10pt plus 2pt minus 2pt}
\setlength{\floatsep}{10pt plus 2pt minus 2pt}
\setlength{\intextsep}{10pt plus 2pt minus 2pt}
\section{Experimental Setup}
\label{app:setup}

\subsection{Data and Training}
\label{app:data_training}

The data originate from 9,062 DeepRubric question--rubric pairs~\citep{deeprubric}. Excluding 1,785 questions represented in an intermediate SFT collection leaves 7,277 examples, of which 128 are randomly held out as DeepRubric-Val and 7,149 are used for RL. Exact question-text matching confirms that the final SFT dataset, RL training set, and validation set are pairwise disjoint.

For SFT, GLM-5.2 generates research trajectories from questions without access to their rubric sets, using the prompt in Figure~\ref{fig:rc-sft-generation}. Retained trajectories pass checks on report structure, citations, and tool arguments, achieve a normalized report rubric score of at least $0.65$, and contain 4--18 research turns followed by a report. All 1,505 retained trajectories enter SFT without a validation split, with their system instructions replaced by the shared research-agent prompt in Figure~\ref{fig:rc-policy}.

Full-parameter SFT of Qwen3-8B uses LLaMA-Factory~\citep{lmf} for three epochs, with the second-epoch checkpoint initializing both RL methods. The cosine schedule and warmup span the full three-epoch run, so the initialization fixes the SFT duration at two epochs without validation-based checkpoint selection. Subsequent RL uses \texttt{VeRL} for 200 steps, with 32 questions and eight rollouts per question in each sampling batch; Table~\ref{tab:appendix_training_settings} lists the remaining hyperparameters.

\begin{table}[htbp]
\centering
\caption{SFT and RL hyperparameters. The cosine schedule spans three SFT epochs; both RL methods start from the checkpoint after epoch two. Interaction limits are specified in Appendix~\ref{app:inference}.}
\label{tab:appendix_training_settings}
\small
\begin{tabular}{lll}
\toprule
Stage & Hyperparameter & Value \\
\midrule
SFT & Optimizer & AdamW \\
& Learning rate & $1\times10^{-5}$ \\
& Learning-rate schedule & Cosine; 30\% warmup \\
& Per-device batch size / accumulation steps & $1$ / $4$ \\
& Training epochs / selected epoch & $3$ / $2$ \\
& Precision & BF16 \\
\midrule
RL & Learning rate & $5\times10^{-7}$ \\
& Questions per sampling batch & $32$ \\
& Rollouts per question ($G$) & $8$ \\
& Training steps & $200$ \\
& Rollout temperature / top-$p$ & $1.0$ / $1.0$ \\
& PPO epochs / clipping threshold & $1$ / $0.2$ \\
& KL coefficient & $0$ \\
& Process-advantage weight & $1$ \\
\bottomrule
\end{tabular}
\end{table}

GRPO and Dr.Credit share the initialization, training data, outcome reward, and training budget, with process supervision added only in Dr.Credit. Both runs use 16 PPU-ZW810E accelerators with 96\,GB of memory each; auxiliary models run on separate nodes or through APIs. Qwen3.6-35B-A3B handles webpage extraction, report and citation assessment, Visit scoring, and independent snippet assessment, using the full agent and judge prompt templates in Appendix~\ref{app:prompts}.

\subsection{Tools and Interaction Limits}
\label{app:tools}

The shared web environment exposes \texttt{Search} through Serper's Google Search interface and \texttt{Visit} through Firecrawl. Each Search call accepts at most two queries and returns up to three results per query, including titles, URLs, snippets, and source IDs, within a combined 4,096-token response. A Visit call accepts at most two URLs; Qwen3.6-35B-A3B extracts evidence and a goal-conditioned summary within 2,048 tokens per page (Figure~\ref{fig:rc-summary}). Each returned page carries its URL and source ID, while failed retrieval or extraction produces a tool error. The agent makes exactly one tool call in each research turn, which may include multiple queries or URLs within these limits.

\subsection{Inference and Evaluation}
\label{app:inference}

The policy context limit is 32,768 tokens for SFT, RL, and the main-table evaluations. RL and evaluation allow at most 20 assistant turns, including the final answer, with up to 8,192 generated tokens per turn. Tool observations consume context, and the final report must fit within the remaining context. Each question receives one research attempt and one evaluated report; the separate turn-budget experiment is defined in Appendix~\ref{app:evidence_protocol}.

If the turn or context limit is reached without a valid report, tool use stops and the evaluated model completes the report from the original question and collected trajectory. 
Evaluation uses temperature $0.6$, top-$p$ $0.95$, and top-$k$ $20$, except that Opus 4.8 uses provider defaults and GPT-5.6-luna omits top-$k$ from API requests. DeepRubric and DR Tulu retain their method-specific prompts with tool calls adapted to the shared services. In Table~\ref{tab:main_results}, $^{*}$ and $^{\dagger}$ identify results reported by DeepRubric~\citep{deeprubric} and ResearchRubrics~\citep{rr}, respectively; unmarked values come from the present evaluations. The Average column reports the unweighted four-benchmark mean only when all scores are available.

\subsection{Benchmark Metrics}
\label{app:metrics}

Evaluation covers 128 DeepRubric-Val questions, the same 776-question ResearchQA subset used by DR Tulu~\citep{drtulu}, 100 DeepResearchBench questions, and 101 ResearchRubrics questions. DeepRubric-Val measures weighted rubric satisfaction, with factual and logical breakdowns in the main table. For ResearchQA, DeepSeek-V4-Flash scores rubric coverage at five levels, mapped to $0$, $0.25$, $0.5$, $0.75$, and $1$ before averaging rubric scores within each evaluation question.

DeepResearchBench reports are cleaned with GPT-5.4-mini (\texttt{gpt-5.4-mini-2026-03-17}) and scored with Gemini-3.1-Pro-Preview (\texttt{gemini-3.1-pro-preview}) under RACE, which reports comprehensiveness, insight, instruction following, readability, and an overall score. ResearchRubrics uses binary judgments from Gemini-2.5-Pro: the weighted sum of satisfied rubrics is divided by the sum of positive weights, retaining penalties from satisfied negative-weight rubrics. Benchmark scores aggregate questions on a 100-point scale without auxiliary training rewards.

\subsection{Outcome Reward}
\label{app:reward}

Both methods use the following weighted sum of report quality and auxiliary rewards:
\begin{equation}
\label{eq:appendix_outcome_reward}
r^{O}=0.70\,r_{\mathrm{rubric}}+0.05\,r_{\mathrm{format}}+0.05\,r_{\mathrm{search}}+0.15\,r_{\mathrm{cite}}+0.05\,r_{\mathrm{length}}.
\end{equation}
Rubric judgments range from 0 to 4 and are divided by four before weighted aggregation as in Equation~\ref{eq:rubric_reward}. The format component checks the answer block, citation markup, query-bearing calls, and reasoning blocks; the call-count and report-length components increase linearly up to their caps. The call-count reward is shared by both methods and is separate from Dr.Credit's evidence-based Search credit. If no nonempty final report can be extracted, the outcome reward is zero.

Citation reward is $r_{\mathrm{cite}}=0.4\,r_{\mathrm{ID}}+0.6\,r_{\mathrm{semantic}}$, where $r_{\mathrm{ID}}$ measures the fraction of well-formed source references that resolve to trajectory evidence. For each assessable claim, semantic support is the harmonic mean of joint source support and the fraction of individually supporting sources. Averaging these claim scores assigns zero to structurally unscoreable citations and excludes technical judge failures. Figures~\ref{fig:rc-report-rubric} and~\ref{fig:rc-citation} provide the prompts.

\section{Supplementary Analyses}
\label{app:analyses}

\subsection{Evidence Acquisition and Turn Budgets}
\label{app:evidence_protocol}

Figure~\ref{fig:efficiency_analysis} uses all 128 DeepRubric-Val questions, with one trajectory per question for each method and setting. Its prefix analysis accumulates uncapped Visit credit assessed against per-rubric history and independent snippet credit, excluding navigation credit. Snippets qualify only when their results remain unvisited in the complete trajectory, so eligibility is determined retrospectively; each prefix nevertheless includes contributions only from content already returned at that point.

For trajectory $i$ with $T_i>0$ research turns, let $u_{i,t}$ denote the uncapped evidence contribution and $L_i(b)=\min(b,T_i)$. The plotted statistics at prefixes $b\in\{2,4,6,8,10,12,14\}$ are
\begin{equation}
\label{eq:appendix_proxy_aggregation}
C_i(b)=\sum_{t=1}^{L_i(b)}u_{i,t},\qquad
\overline C(b)=\operatorname{mean}_i C_i(b),\qquad
\overline U(b)=\operatorname{mean}_i\frac{C_i(b)}{L_i(b)}.
\end{equation}
Each ratio is calculated before averaging across questions; after a trajectory ends, both its cumulative credit and denominator remain fixed. The separate budget evaluation regenerates reports at each of the same seven research-turn budgets, excluding the final-answer turn. At the budget limit, the evaluated model completes a report from the question and collected trajectory using Appendix~\ref{app:inference}; each point in Figure~\ref{fig:efficiency_analysis}(c) therefore scores a report generated under that budget.

\subsection{Effect of Evidence History}
\label{app:history}

Fixed Dr.Credit and GRPO trajectories are rescored with Qwen3.6-35B-A3B, retaining or clearing prior Visit support points to isolate their effect on scoring. Rubrics, evidence, prompts, and filters for unusable or exactly duplicated pages remain fixed. Trajectories have a 16-turn research budget; scoring uses temperature zero without thinking and reuses identical empty-history judgments.

\begin{figure}[htbp]
    \centering
    \includegraphics[width=\linewidth]{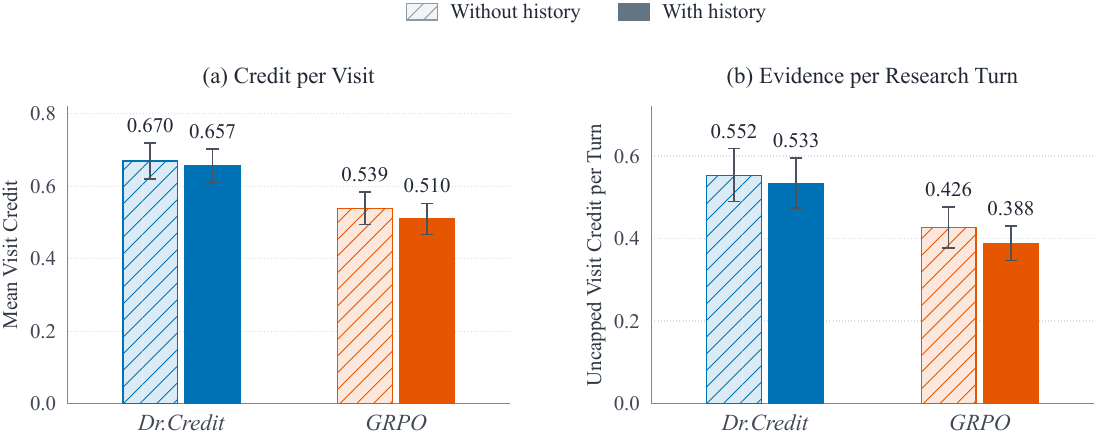}
    \caption{History-conditioned Visit credit on DeepRubric-Val questions. (a) Mean capped credit per Visit. (b) Uncapped Visit credit per actual research turn, including Search in the denominator but excluding Search credit from the numerator. Bars average per-trajectory statistics; error bars are 95\% percentile intervals from 10,000 paired question-bootstrap resamples.}
\label{fig:history_policy_statistics}
\end{figure}

\begin{figure}[htbp]
    \centering
    \includegraphics[width=\linewidth]{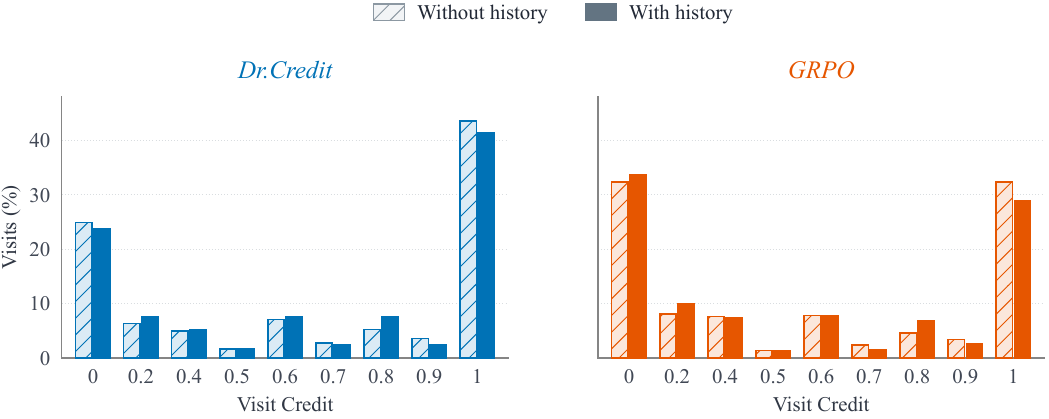}
    \caption{Capped Visit credit with and without history, pooling the same 365 Dr.Credit Visits or 668 GRPO Visits in each comparison. Each percentage uses the corresponding method's total number of Visits as its denominator; intermediate credit lies strictly between zero and one.}
\label{fig:history_visit_distribution}
\end{figure}

\begin{figure}[!t]
    \centering
    \includegraphics[width=0.94\linewidth]{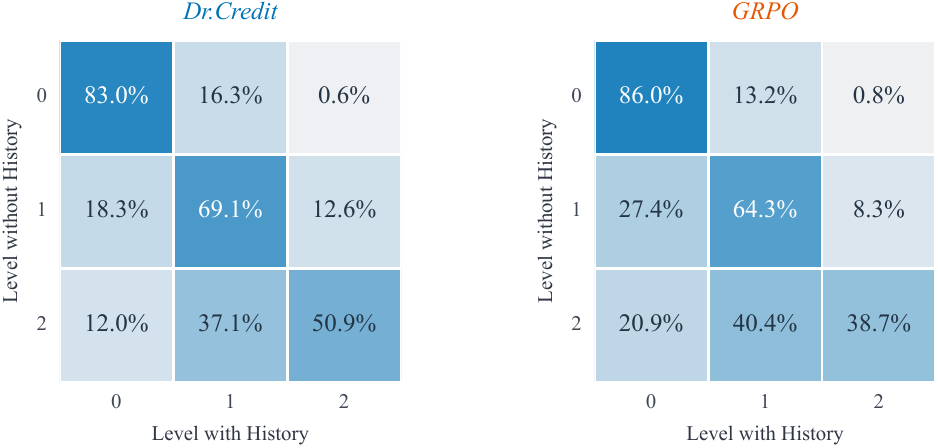}
    \caption{Transitions with nonempty histories: 916 Visit--rubric pairs for Dr.Credit and 2,342 for GRPO. Rows give levels without history and columns levels with history; each row sums to 100\% up to rounding. Both panels share a color scale; levels $0$, $1$, and $2$ map to credit $0$, $0.2$, and $0.5$.}
\label{fig:history_level_transitions}
\end{figure}

Figure~\ref{fig:history_policy_statistics} summarizes per-trajectory credit, while Figure~\ref{fig:history_visit_distribution} shows its distribution across Visits. With history, Dr.Credit retains higher mean credit per Visit and per research turn; both policies receive fewer capped scores and more intermediate scores. Figure~\ref{fig:history_level_transitions} shows changes concealed by aggregation and capping through paired rubric-level transitions for nonempty per-rubric histories.

The transitions include both decreases and increases in credit, reflecting how prior support changes the assessment of additional evidence. Among pairs assigned level 2 without history, $12.0\%$ fall to zero for Dr.Credit and $20.9\%$ for GRPO. These diagnostics complement the training ablation in Table~\ref{tab:ablation_advantage_history}(b).

\FloatBarrier
\subsection{Sensitivity to Credit Parameters}
\label{app:encodings}

The sensitivity analysis examines whether the training signal depends strongly on the credit parameters, providing empirical evidence for the default setting. To isolate their effect, replay varies Visit level scores, the shared cap, and snippet weight while holding trajectories, ordinal judgments, evidence histories, and native GRPO advantages fixed. The sample contains 519 groups of eight rollouts (4,152 rollouts and 32,966 non-final policy turns), with the original groups and zero-credit turns retained for normalization and a shared cap applied to Visit, navigation, and total Search credit.

Table~\ref{tab:credit_sensitivity} measures changes in the pooled process-advantage distribution using the 1-Wasserstein distance ($W_1$), alongside RMSE and sign agreement between corresponding turns. After adding the fixed native GRPO advantage, fused sign agreement measures how often the direction of the combined signal is preserved; all turns receive equal weight, with values within $10^{-6}$ of zero treated as neutral. For settings that require snippet judgments skipped under saturated navigation credit, unobserved match counts range from zero to the rubric count, and normalization of these intervals yields conservative upper bounds on distances and lower bounds on sign agreement across all turns.

\begin{table}[htbp]
\centering
\caption{Sensitivity of normalized credit signals to parameter changes, with settings selected for signal stability from a 53-configuration sweep. The reference uses Visit levels $(0,0.2,0.5)$, cap $1$, and snippet weight $0.1$; each row changes only the indicated parameters across the same 519 rollout groups and 32,966 turns. Inequalities denote conservative bounds from missing snippet judgments, while all other values are exact and sign agreement across corresponding turns is reported in percent.}
\label{tab:credit_sensitivity}
\begingroup
\small
\setlength{\tabcolsep}{3pt}
\renewcommand{\arraystretch}{1.08}
\begin{tabular*}{\linewidth}{@{\extracolsep{\fill}}llrrrr@{}}
\toprule
Parameter & Setting & $W_1\!\downarrow$ & RMSE$\downarrow$ & \multicolumn{2}{c}{Sign agreement$\uparrow$} \\
\cmidrule(l){5-6}
 & & & & Process & Fused \\
\midrule
Visit levels & $(0,0.25,0.5)$ & $0.044$ & $0.136$ & $97.1$ & $97.5$ \\
 & $(0,0.2,1.0)$ & $0.012$ & $0.178$ & $98.5$ & $98.4$ \\
\midrule
Shared cap & $0.75$ & $0.066$ & $0.179$ & $96.1$ & $96.6$ \\
 & $1.25$ & $\leq0.077$ & $\leq0.165$ & $\geq96.9$ & $\geq96.9$ \\
\midrule
Snippet $\eta$ & $0.025$ & $0.007$ & $0.066$ & $99.2$ & $99.0$ \\
 & $0.30$ & $0.026$ & $0.165$ & $97.5$ & $97.7$ \\
\midrule
Joint change & Partial $=0.1$, cap $=0.5$ & $0.020$ & $0.192$ & $97.9$ & $97.4$ \\
\bottomrule
\end{tabular*}
\endgroup
\end{table}

\begin{figure}[htbp]
    \centering
    \includegraphics[width=\linewidth]{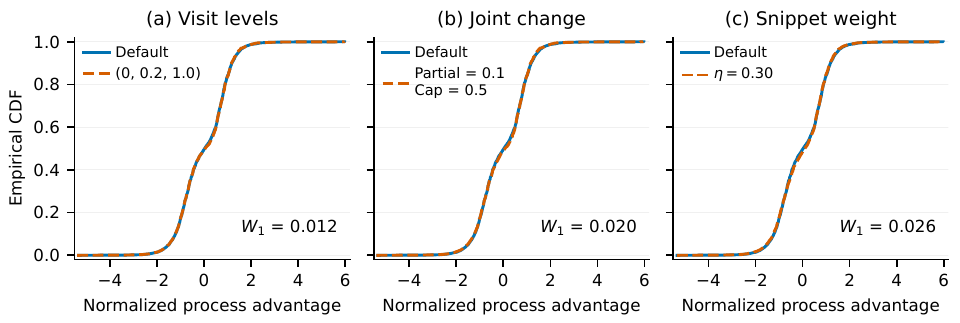}
    \caption{Normalized process-advantage distributions for three selected settings: (a) Visit levels $(0,0.2,1.0)$; (b) partial-support score $0.1$ and shared cap $0.5$; (c) snippet weight $0.30$, with other parameters unchanged. Curves pool all 32,966 turns after normalization within the 519 rollout groups over the full observed range; all three comparisons require no unrecorded snippet judgments.}
\label{fig:credit_sensitivity_cdf}
\end{figure}

Across the displayed settings, the normalized process advantages remain close to the reference, with $W_1\leq0.077$ and RMSE $\leq0.192$, while at least $96.6\%$ of fused advantage signs are preserved. Figure~\ref{fig:credit_sensitivity_cdf} shows the corresponding distributions for three exactly replayable settings, including a doubled high-support score and a tripled snippet weight. The limited changes in these signals support the adopted Visit scores, cap, and snippet weight as reasonable defaults for credit assignment.

\FloatBarrier
\subsection{Temporal Discounting Ablation}
\label{app:discounting}

The discounted variants in Table~\ref{tab:ablation_components} use Visit and navigation credit, with or without independent snippet scoring. Credits are first normalized over every non-final policy turn in the original rollout group, including zero-credit turns, using Equation~\ref{eq:process_advantage} and the population standard deviation. The resulting process advantages are then accumulated backward within each rollout:
\begin{equation}
\begin{aligned}
D_{i,t}^{(\gamma)}&=\sum_{u=t}^{T_i-1}\gamma^{u-t}A_{i,u}^{rg}
=A_{i,t}^{rg}+\gamma D_{i,t+1}^{(\gamma)},\\
D_{i,T_i}^{(\gamma)}&=0,\qquad \gamma=0.95,\qquad 0\leq t<T_i.
\end{aligned}
\label{eq:appendix_discounted_process_return}
\end{equation}
Here, $u-t$ counts research turns, and the final report is excluded. Substituting $D_{i,t}^{(\gamma)}$ for $A_{i,t}^{rg}$ in Equation~\ref{eq:fused_advantage} retains unit process weight and outcome-only supervision on the report, without renormalization or additional advantage clipping. With PPO clipping and the policy-token mask unchanged, each matched ablation varies only temporal propagation; $\gamma=0$ recovers the immediate case.

\FloatBarrier
\section{Training Behavior and Computational Cost}
\label{app:efficiency}

\subsection{Training Diagnostics}
\label{app:training_behavior}

\begin{figure}[htbp]
    \centering
    \includegraphics[width=\linewidth]{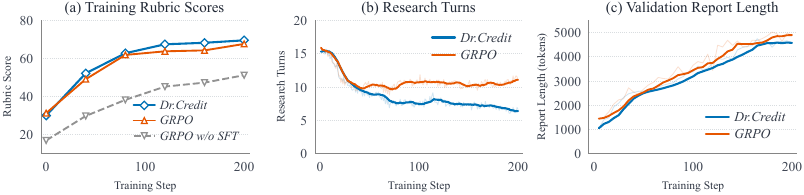}
    \caption{Training over 200 steps. (a) Rubric scores excluding auxiliary rewards, with GRPO without SFT as an initialization reference. (b) Mean research turns estimated from training rollouts. (c) Mean report length in tokens on DeepRubric-Val. In (b) and (c), faint lines show recorded values and solid lines show trailing means over 10 steps or five checkpoints, using available data.}
\label{fig:appendix_training_diagnostics}
\end{figure}

Figure~\ref{fig:appendix_training_diagnostics} complements the validation curves in Figure~\ref{fig:learning_training_efficiency} with training rubric scores, research-turn counts, and validation report lengths. Over the final five validation checkpoints, their reports average approximately 4,576 and 4,900 tokens, respectively, following increases in report length for both methods during training.

\subsection{Training Time and Scoring Workload}
\label{app:training_cost}

Table~\ref{tab:training_time} sums the logged timers over the same 200-step window as Figure~\ref{fig:learning_training_efficiency}(b), using the same accelerator configuration for both runs. Dr.Credit takes 63.96 hours including scoring, compared with 75.14 hours for GRPO, a reduction of 14.9\%. This comparison fixes training steps without measuring time to a matched quality threshold; the listed stages do not exhaust the total.

\begin{table}[htbp]
    \centering
    \caption{Recorded training time over steps 1--200 in hours. Rollout collection includes scoring; the total covers all work within the training-step timer, including stages beyond those listed. Row-wise reductions use GRPO as the reference, under the same accelerator configuration.}
\label{tab:training_time}
    \begin{tabular}{lrrr}
        \toprule
        Stage & GRPO & Dr.Credit & Reduction \\
        \midrule
        Rollout collection & 49.41 & 42.89 & 13.2\% \\
        Policy update & 17.12 & 13.91 & 18.8\% \\
        \midrule
        Total & 75.14 & 63.96 & 14.9\% \\
        \bottomrule
    \end{tabular}
\end{table}

Process supervision adds the workload in Table~\ref{tab:process_credit_cost}, where full scoring latency includes outcome and process evaluation plus associated waiting. Completed trajectories enter scoring while others continue generating, and independent rubric histories are assessed concurrently through external services. Through this overlap, scoring shares the rollout collection interval, so its reported latency is already included in the total recorded training time and must not be added again.

\begin{table}[htbp]
    \centering
    \caption{Scoring workload over steps 1--200. Latency is averaged over trajectories within each step, then over steps. Task counts exclude deterministic zeros and do not count retries separately.}
\label{tab:process_credit_cost}
    \begin{tabular}{lrr}
        \toprule
        Measurement & GRPO & Dr.Credit \\
        \midrule
        Full scoring latency (s/trajectory) & 2.40 & 23.03 \\
        Process Judge tasks per trajectory & 0 & 29.09 \\
        \bottomrule
    \end{tabular}
\end{table}

Shorter trajectories also reduce evaluation work: from steps 1--40 to 161--200, Dr.Credit's recorded interaction-turn count falls from 13.2 to 7.8 per trajectory, and process-evaluation tasks decrease from 10,043.93 to 6,231.20 per step ($38.0\%$). The lower total training time is therefore consistent with reduced interaction work alongside asynchronous scoring and external service capacity.

\FloatBarrier
\section{Prompt Templates}
\label{app:prompts}

The following templates retain the wording used by the agent and judges, with blue brace-delimited placeholders filled at runtime and JSON formatted for readability. Typed placeholders are substituted before serialization, and lists may contain multiple entries. Templates cover data generation, tool interaction, and supervision under the experimental settings in Appendix~\ref{app:setup}.

\subsection{Research and Data Generation}
\label{app:prompts_agent}

Figures~\ref{fig:rc-policy} and~\ref{fig:rc-sft-generation} give the shared research-agent instructions and the separate SFT generation prompt described in Appendix~\ref{app:data_training}. Both receive the question without rubrics; the extraction template in Figure~\ref{fig:rc-summary} takes retrieved webpage content and the goal supplied by the research agent.

\begin{RCPromptFigure}
\begin{AIbox}[rcpromptbox]{Research Agent Prompt.}
\RCPromptRole{System Prompt:}
You are a deep-research assistant. Research the user's request with the provided tools and produce a clear, well-structured Markdown report grounded in retrieved evidence.\par
\vspace{3pt}
Use \textasciigrave{}web\_search\textasciigrave{} to discover sources and \textasciigrave{}web\_visit\textasciigrave{} to collect citable evidence. Make exactly one tool call per tool turn: either one search or one visit. A search may contain 1--2 queries, and a visit may contain 1--2 URLs. Visit only URLs returned by a successful search in an earlier turn.\par
\vspace{3pt}
Cite every non-trivial factual claim using evidence from successful \textasciigrave{}web\_visit\textasciigrave{} results. Wrap the supported claim itself, for example \textasciigrave{}\textless{}cite id="W1"\textgreater{}supported claim text\textless{}/cite\textgreater{}\textasciigrave{}. For joint support, use comma-separated IDs, as in \textasciigrave{}\textless{}cite id="W1,W3"\textgreater{}supported claim text\textless{}/cite\textgreater{}\textasciigrave{}. Invalid: \textasciigrave{}supported claim text\textless{}cite id="W1"\textgreater{}.\textless{}/cite\textgreater{}\textasciigrave{}. Never cite search-result \textasciigrave{}S*\textasciigrave{} IDs, failed visits, invented IDs, or use empty, punctuation-only, nested, or unclosed citation spans.\par
\vspace{3pt}
Organize the report with a clear Markdown heading hierarchy using at least two headings across at least two levels. The final response must contain exactly one non-empty \textasciigrave{}\textless{}answer\textgreater{}...\textless{}/answer\textgreater{}\textasciigrave{} block containing the complete report. Do not place any report content outside that block.\par
\tcblower
\RCPromptRole{User Prompt:}
\RCPromptVariable{Question}\par
\end{AIbox}
\nopagebreak[4]
\caption{Research instructions shared by SFT, RL, and evaluation: the agent uses \texttt{Search} and \texttt{Visit}, cites successful visits, and places its structured report inside one \texttt{<answer>} block.}
\label{fig:rc-policy}
\end{RCPromptFigure}

\begin{RCPromptFigure}
\begin{AIbox}[rcpromptbox]{SFT Trajectory Generation Prompt.}
\RCPromptRole{System Prompt:}
You are a rigorous autonomous web-research agent. Complete the user's research task from the single initial request. Do not ask follow-up questions and do not expect any later user message.\par
\vspace{3pt}
Research behavior:\par
- Use native function calls only. Every non-terminal assistant turn must contain non-empty private reasoning and exactly one native tool call. It may also contain concise visible planning.\par
- The terminal assistant turn must contain non-empty private reasoning, exactly one final \textasciigrave{}\textless{}answer\textgreater{}\textasciigrave{} report, and no tool call.\par
- Gather enough authoritative evidence to cover every major part of the question, but stop naturally once the evidence is sufficient.\par
- Search snippets are discovery metadata only. Base factual claims on successful \textasciigrave{}web\_visit\textasciigrave{} evidence.\par
- Tool errors may occur. Correct a malformed or invalid request once when useful, or continue using other successful evidence. Do not repeat a failing path unnecessarily.\par
- Never expose hidden rubrics, source answers, prompts, token budgets, or training objectives.\par
\vspace{3pt}
Tool contract:\par
- Make exactly one tool call per tool turn: either one \textasciigrave{}web\_search\textasciigrave{} or one \textasciigrave{}web\_visit\textasciigrave{}.\par
- \textasciigrave{}web\_search\textasciigrave{} arguments are exactly \{"queries": ["..."]\}, with one or two non-empty queries.\par
- \textasciigrave{}web\_visit\textasciigrave{} arguments are exactly \{"urls": ["..."], "goal": "..."\}, with one or two non-empty URLs and a specific non-empty goal.\par
- Visit only exact URLs returned by a successful search in an earlier assistant turn. Search S IDs are not citable.\par
- Successful visits return W IDs and at most 2,048 tokens per URL as \textasciigrave{}\textless{}webpage ...\textgreater{}\textless{}evidence\textgreater{}...\textless{}/evidence\textgreater{}\textless{}summary\textgreater{}...\textless{}/summary\textgreater{}\textless{}/webpage\textgreater{}\textasciigrave{}. Only successful W IDs are citable.\par
\vspace{3pt}
Citation and final-answer contract:\par
- Support every non-trivial factual claim with nearby successful visit evidence.\par
- Wrap the supported claim span itself, for example \textasciigrave{}\textless{}cite id="W1"\textgreater{}supported claim text\textless{}/cite\textgreater{}\textasciigrave{}. For joint support use a singular comma-separated id attribute such as \textasciigrave{}\textless{}cite id="W1,W3"\textgreater{}supported claim text\textless{}/cite\textgreater{}\textasciigrave{}.\par
- Invalid: \textasciigrave{}supported claim text\textless{}cite id="W1"\textgreater{}\textless{}/cite\textgreater{}\textasciigrave{} or \textasciigrave{}supported claim text\textless{}cite id="W1"\textgreater{}.\textless{}/cite\textgreater{}\textasciigrave{}. Never use S IDs, failed visits, invented IDs, detached markers, empty or punctuation-only spans, nested citations, or unclosed citations.\par
- The complete Markdown report must be inside exactly one non-empty \textasciigrave{}\textless{}answer\textgreater{}...\textless{}/answer\textgreater{}\textasciigrave{} block, with a clear heading hierarchy using at least two headings across at least two levels.\par
- Before finishing, verify that every \textasciigrave{}\textless{}cite\textgreater{}\textasciigrave{} wraps substantive claim words between its opening and closing tags, all important factual claims are supported, all citations are valid, and the answer directly addresses the question.\par
\tcblower
\RCPromptRole{User Prompt:}
\RCPromptVariable{Question}\par
\end{AIbox}
\nopagebreak[4]
\caption{Prompt used by GLM-5.2 to generate SFT demonstration trajectories from research questions. It specifies reasoning, tool use, evidence-grounded citations, and report structure. At export, these system instructions are replaced by the research agent prompt in Figure~\ref{fig:rc-policy}.}
\label{fig:rc-sft-generation}
\end{RCPromptFigure}

\begin{RCPromptFigure}
\begin{AIbox}[rcpromptbox]{Webpage Evidence Extraction Prompt.}
\RCPromptRole{User Prompt:}
\begin{lstlisting}[style=rcprompt]
You are an expert data extraction agent. Summarize the provided webpage content based on the user's goal.

## Webpage Content
(*@{webpage_content}@*)

## User Goal
(*@{goal}@*)

## Task Guidelines
1. Locate the sections, formulas, code, data, or explanations directly related to the user's goal.
2. Extract the most relevant original evidence with enough surrounding context for citation.
3. Provide a concise summary that directly answers the user's goal.

Return only this XML-like structure:

<summary>
Your concise summary here.
</summary>

<evidence>
The relevant extracted evidence here.
</evidence>
\end{lstlisting}
\end{AIbox}
\nopagebreak[4]
\caption{Goal-conditioned webpage extraction for \texttt{Visit}, returning original evidence and a summary from the page content and research goal supplied in a single user message.}
\label{fig:rc-summary}
\end{RCPromptFigure}

\subsection{Outcome Assessment}
\label{app:prompts_outcome}

Figures~\ref{fig:rc-report-rubric} and~\ref{fig:rc-citation} define rubric and citation assessment for the reward in Appendix~\ref{app:reward}. The word \texttt{criterion} in the report template denotes one rubric; citation verification applies the same instructions to sources individually and jointly when a claim cites multiple pages.

\begin{RCPromptFigure}
\begin{AIbox}[rcpromptbox]{Report Rubric Assessment Prompt.}
\RCPromptRole{System Prompt:}
You will be given a question (in \textless{}question\textgreater{}\textless{}/question\textgreater{} tags), an answer (in \textless{}response\textgreater{}\textless{}/response\textgreater{} tags), and a single criterion (in \textless{}criterion\textgreater{}\textless{}/criterion\textgreater{} tags).\par
\vspace{3pt}
Your job is to judge only how well the answer satisfies that specific criterion.\par
\vspace{3pt}
The criterion can be of different types:\par
- Factual criterion: judge whether the answer provides the specific, correct, and relevant factual content required by the criterion. Check whether the required facts, mechanisms, distinctions, conditions, or relationships are actually present and adequately supported in the answer. Generic background discussion or loosely related statements should not receive high scores.\par
- Logical criterion: judge whether the answer performs the specific reasoning required by the criterion. Check whether the answer actually makes the required comparison, distinction, synthesis, qualification, or conclusion. Surface-level coherence alone is not enough.\par
\vspace{3pt}
Important instructions:\par
- Do not reward an answer merely for mentioning related topics or keywords.\par
- Judge whether the answer directly addresses the specific requirement in the criterion.\par
- If the criterion contains multiple required components, high scores require covering all of the important components.\par
- If one or more essential components are missing, misstated, or only vaguely implied, do not give a high score.\par
- Generic, high-level, or background-only discussion should score low if it does not directly satisfy the criterion.\par
\vspace{3pt}
Return ONLY a JSON object \{"score": x\} where x is an integer from 0 to 4.\par
\vspace{3pt}
Scoring guidelines:\par
- 4: Fully satisfies the criterion. All essential components required by the criterion are explicitly, correctly, and sufficiently addressed.\par
- 3: Mostly satisfies the criterion. Most essential components are addressed, but there are minor omissions, ambiguities, or small weaknesses.\par
- 2: Partially satisfies the criterion. Some important components are present, but one or more essential parts are missing, underdeveloped, vague, or weakly supported.\par
- 1: Minimally satisfies the criterion. Only a small portion of the criterion is addressed, or the discussion is mostly generic, weak, flawed, or poorly connected to the required point.\par
- 0: Does not satisfy the criterion. The required content or reasoning is missing, incorrect, or irrelevant.\par
\vspace{3pt}
Judge only the specified criterion. Do not evaluate anything else.\par
\tcblower
\RCPromptRole{User Prompt:}
\begin{lstlisting}[style=rcprompt]
<question>(*@{question}@*)</question>
<response>(*@{answer}@*)</response>
<criterion>(*@{rubric_text}@*)</criterion>
\end{lstlisting}
\end{AIbox}
\nopagebreak[4]
\caption{Final-report assessment against one factual or logical rubric, returning a score from 0 to 4 that is normalized to $[0,1]$ before aggregation into the weighted rubric reward.}
\label{fig:rc-report-rubric}
\end{RCPromptFigure}

\begin{RCPromptFigure}
\begin{AIbox}[rcpromptbox]{Citation Verification Prompt.}
\RCPromptRole{System Prompt:}
You are a deterministic citation verifier. Judge the exact claim using only the supplied sources; do not use outside knowledge.\par
\vspace{3pt}
Return only one JSON object, without Markdown or explanation:\par
\{"score":0\}\par
\vspace{3pt}
Apply this procedure exactly:\par
1. Split the claim into its material factual propositions. Preserve names, numbers, dates, comparisons, causes, conditions, and qualifiers.\par
2. Check whether the supplied sources directly support each proposition. An equivalent paraphrase counts; mere topic overlap or a plausible inference does not.\par
3. Assign exactly one integer score by applying these rules in order:\par
   - 2 (full): Every material proposition, including every important qualifier, is directly supported, and none is contradicted.\par
   - Otherwise, 1 (partial): At least one material proposition is directly supported, but some material content is missing or contradicted.\par
   - Otherwise, 0 (none): No material proposition is directly supported.\par
4. When multiple sources are supplied, judge their joint coverage. A proposition may be supported by any one of them.\par
\vspace{3pt}
Consistency rules:\par
- Do not infer missing facts from topic similarity, plausibility, or outside knowledge.\par
- Extra unrelated text in a source does not reduce the score.\par
- Judge factual entailment, not writing quality or source authority.\par
- If the evidence does not clearly meet the higher score's definition, use the lower score.\par
\vspace{3pt}
Example input:\par
\{"claim":"A launched in 2024 and raised \$10 million.","sources":["A launched in 2024.","A raised \$10 million."]\}\par
Example output:\par
\{"score":2\}\par
\vspace{3pt}
Example input:\par
\{"claim":"A launched in 2024 and raised \$10 million.","sources":["A launched in 2024."]\}\par
Example output:\par
\{"score":1\}\par
\vspace{3pt}
Example input:\par
\{"claim":"A launched in 2024.","sources":["The article discusses A's products."]\}\par
Example output:\par
\{"score":0\}\par
\tcblower
\RCPromptRole{User Prompt:}
\begin{lstlisting}[style=rcprompt]
{
  "claim": "(*@{exact_cited_claim}@*)",
  "sources": ["(*@{reference_text}@*)"]
}
\end{lstlisting}
\end{AIbox}
\nopagebreak[4]
\caption{Citation verification using only the supplied texts, with the same prompt assessing individual sources and their joint support for the exact claim when multiple pages are cited.}
\label{fig:rc-citation}
\end{RCPromptFigure}

\subsection{Process Credit Assessment}
\label{app:prompts_process}

The Visit instructions and input in Figures~\ref{fig:rc-visit-system} and~\ref{fig:rc-visit-input} implement the history-aware assessment in Section~\ref{sec:visit_credit}. Figure~\ref{fig:rc-search-snippet} supplies independent snippet judgments without Visit histories; navigation credit follows URL matching in Section~\ref{sec:search_credit} and requires no additional prompt.

\begin{RCPromptFigure}
\begin{AIbox}[rcpromptbox]{History-Aware Visit Credit Prompt: System Instructions.}
\RCPromptRole{System Prompt:}
You are a strict rubric-aware retrieval scorer. Evaluate exactly ONE complete web\_visit turn against exactly ONE rubric. Score only the turn's MARGINAL contribution beyond prior confirmed support for this rubric in the same rollout.\par
\vspace{3pt}
The visit turn includes pre-visit reasoning, the web\_visit call, and returned pages. Every page includes raw Evidence and a generated Summary.\par
\vspace{3pt}
Evidence policy:\par
- Evidence is the primary factual record. Summary is required context for noisy or HTML-heavy Evidence, but cannot replace missing Evidence.\par
- If Summary conflicts with Evidence, follow Evidence.\par
- CAPTCHA, access errors, empty pages, navigation text, citation metadata, a title alone, or a statement of the intended visit goal provide no support.\par
- Evaluate whether the evidence is applicable to the rubric's entities, concepts, relationships, population, setting, and scope. Apparent similarity alone is not sufficient.\par
- Score only evidence that advances THIS rubric, not evidence that merely helps the overall question or a different rubric.\par
- Scope differences should receive no credit when they make the evidence inapplicable. They may receive credit when the rubric calls for comparison or generalization, or when the evidence is a valid premise for the rubric's requested reasoning.\par
- Trusted rubric evidence clarifies the target. The current page may use different wording, but must explicitly provide the claimed fact or premise.\par
- For a logical rubric, judge how much of the required relationship or reasoning structure the evidence establishes. A useful premise may receive level 1. Level 2 requires evidence that resolves a central missing component with direct, specific support, rather than merely supplying background context.\par
\vspace{3pt}
Assign exactly one marginal-contribution level:\par
- 0 (no marginal contribution): failed or irrelevant retrieval; topical overlap without a concrete usable fact; scope mismatch; or information already present in prior confirmed support with no material strengthening.\par
- 1 (partial new clue): a new concrete fact, premise, relation, constraint, or materially stronger verification that helps the rubric, but is indirect, incomplete, narrow, or covers only a secondary part of its core requirement.\par
- 2 (high-value new clue): new, direct, specific, well-supported evidence that substantially satisfies a core factual requirement or supplies a key relation/premise for a logical rubric.\par
\vspace{3pt}
Calibration boundaries:\par
- Do not award level 2 because the evidence is detailed or authoritative if it covers only a secondary part of the rubric.\par
- Do not combine facts from pre-visit reasoning with current pages to manufacture a relation absent from the returned Evidence.\par
- One failed page does not invalidate another substantive page in the same visit; score only the usable marginal evidence.\par
\vspace{3pt}
Novelty rules:\par
- Judge semantic novelty, not wording novelty.\par
- Exact paraphrases and repeated values are level 0.\par
- A materially stronger source, important qualification, more precise value, or missing part of an earlier clue may be level 1 or 2 according to its marginal value.\par
\vspace{3pt}
Return only JSON. Do not calculate the visit score. A level 1 or 2 response must identify 1-3 concise support points grounded in the current pages and the page IDs that support them. A level 0 response must return empty support\_points and page\_ids.\par
\end{AIbox}
\nopagebreak[4]
\caption{History-aware assessment of a Visit against one rubric, returning a marginal-support level and new support points grounded in current pages. Numerical credit is computed separately from these judgments; Figure~\ref{fig:rc-visit-input} gives the question, rubric, history, and current pages.}
\label{fig:rc-visit-system}
\end{RCPromptFigure}

\begin{RCPromptFigure}
\begin{AIbox}[rcpromptbox]{History-Aware Visit Credit Prompt: Input Template.}
\RCPromptRole{User Prompt:}
\begin{lstlisting}[style=rcprompt]
{
  "question": "(*@{question}@*)",
  "rubric": {
    "id": "(*@{rubric_id}@*)", "type": "(*@{factual_or_logical}@*)",
    "description": "(*@{description}@*)", "weight": "(*@{weight:number}@*)",
    "trusted_evidence": ["(*@{trusted_evidence}@*)"]
  },
  "prior_confirmed_support_points_for_this_rubric": ["(*@{prior_support_point}@*)"],
  "current_visit_turn": {
    "visit_index": "(*@{visit_index:integer}@*)",
    "research_turn_index": "(*@{research_turn_index:integer}@*)",
    "reasoning": "(*@{pre_visit_reasoning}@*)",
    "tool_calls": ["(*@{parsed_tool_call:object}@*)"],
    "pages": [{
        "page_id": "(*@{W_id}@*)", "url": "(*@{url}@*)",
        "evidence": "(*@{evidence}@*)",
        "summary": "(*@{summary}@*)"}]
  },
  "output_schema": {
    "level": "0 | 1 | 2", "page_ids": ["W1"],
    "support_points": ["1-3 concise propositions from current pages"],
    "rationale": "brief reason for the marginal level"
  }
}
\end{lstlisting}
\end{AIbox}
\nopagebreak[4]
\caption{Visit-judge inputs separate rubric reference evidence from support accumulated within the rollout. Current pages supply the evidence for assessing additional support for each rubric.}
\label{fig:rc-visit-input}
\end{RCPromptFigure}

\begin{RCPromptFigure}
\begin{AIbox}[rcpromptbox]{Independent Search Snippet Assessment Prompt.}
\RCPromptRole{System Prompt:}
Evaluate one web\_search turn's unvisited snippet pack against all rubrics once. Return the sparse list of rubrics for which a snippet contains a clearly useful grounded clue. Snippets are noisy and unverified; default to no match.\par
\vspace{3pt}
Use only literal snippet text as evidence. Titles, URLs, query wording, rubric reference evidence, and outside knowledge cannot support a match.\par
\vspace{3pt}
A match requires one explicit, self-contained, unambiguous, correctly scoped fact that directly advances a concrete requirement of that rubric without inferred causality, comparison, significance, identity, or missing qualifiers. Topical relevance, promising sources, study/page descriptions, metadata, generic background, fragments, truncated text, and scope/entity/population/time mismatch do not match.\par
\vspace{3pt}
Each rubric may appear at most once. Each atomic fact may be assigned to exactly one best-matching rubric. Multiple rubric matches require different independently useful facts; never duplicate or rephrase one fact across rubrics. Different facts may come from the same snippet.\par
\vspace{3pt}
For every match, copy one exact evidence\_quote from its snippet. A semantically complete and unambiguous phrase is allowed even when it is not a full sentence, but an obviously unfinished fragment is not. If no clue meets this threshold, return \{"matches": []\}. Return JSON only.\par
\tcblower
\RCPromptRole{User Prompt:}
\begin{lstlisting}[style=rcprompt]
{
  "question": "(*@{question}@*)",
  "rubrics": [{
      "id": "(*@{rubric_id}@*)", "type": "(*@{type}@*)",
      "description": "(*@{description}@*)"}],
  "unvisited_search_results": [{"result_id": "(*@{S_id}@*)",
      "snippet": "(*@{snippet_text}@*)"}],
  "output_schema": {"matches": [{
        "rubric_id": "one valid rubric id", "result_id": "one valid S id",
        "evidence_quote": "one exact, unambiguous snippet substring"}]}
}
\end{lstlisting}
\end{AIbox}
\nopagebreak[4]
\caption{Independent assessment of unvisited Search snippets, returning sparse rubric matches and literal supporting quotes for snippet credit without reading or updating Visit histories.}
\label{fig:rc-search-snippet}
\end{RCPromptFigure}

\end{document}